\ifdefined\XeTeXversion\else\pdfoutput=1\fi 
\documentclass[10pt]{article}

\usepackage[letterpaper,textwidth=5.5in,textheight=9in,headheight=12pt,headsep=25pt,footskip=30pt]{geometry}
\usepackage[T1]{fontenc}
\usepackage[utf8]{inputenc}
\usepackage{times}
\usepackage{microtype}
\usepackage{amsmath,amssymb,amsthm}
\usepackage{booktabs}
\usepackage{multirow}
\usepackage{graphicx}
\usepackage{xcolor}
\usepackage{tikz}
\usetikzlibrary{arrows.meta,positioning,calc}
\usepackage[round,sort]{natbib}
\usepackage[font=small,labelfont=bf]{caption}
\usepackage{subcaption}
\usepackage{enumitem}
\PassOptionsToPackage{hyphens}{url}\usepackage{url}
\usepackage{float}
\usepackage[colorlinks=true,linkcolor=blue!60!black,citecolor=blue!60!black,urlcolor=blue!60!black]{hyperref}
\setlist{nosep,leftmargin=*}
\usepackage{titlesec}
\titleformat{\section}{\large\bfseries\raggedright}{\thesection}{1em}{}
\titleformat{\subsection}{\normalsize\bfseries\raggedright}{\thesubsection}{1em}{}
\titleformat{\paragraph}[runin]{\normalsize\bfseries}{}{0pt}{}[.]
\titlespacing*{\section}{0pt}{2.0ex plus 0.5ex minus 0.2ex}{1.2ex plus 0.3ex}
\titlespacing*{\subsection}{0pt}{1.6ex plus 0.5ex minus 0.2ex}{0.8ex plus 0.3ex}
\titlespacing*{\paragraph}{0pt}{1.0ex plus 0.4ex}{0.6em}

\newcommand{\dtp}{DTP}
\newcommand{\KL}{\mathrm{KL}}
\newcommand{\fo}{\textsc{fo}}
\newcommand{\addsc}{\textsc{add}}
\newcommand{\qwen}{Qwen3-0.6B}
\newcommand{\danube}{Danube3-500M}

\title{\LARGE\bfseries Affinity-Aware Sharding for Delayed Tensor Parallelism}

\author{%
  Eloi de Reynal\\
  \texttt{eloidereynal@gmail.com}
}
\date{}

\begin{document}
\maketitle

\begin{abstract}
Delayed Tensor Parallelism (\dtp) removes the blocking all-reduce of tensor-parallel Transformer inference. Every device adds its own partial output to its residual stream (and broadcasts it) immediately, but only gathers (receives) the other devices' partials $\delta$ modules later. This architecture forces some degree of independence between devices, as the communication \emph{delay} also degrades its \emph{quality}: each partial is computed from a stream that lacks the other devices' latest writes. A TP to \dtp{} change therefore amounts to a real architecture change, and dense Transformer models need to be retrained or distilled after adaptation. We show that \dtp{} breaks the permutation symmetry of neurons inside FFNs and of KV heads inside attention modules, and that this symmetry breakage makes the sharding itself a modelling decision. We show that maximising the affinity between the KV heads and the FFN neurons co-located on a device, by permuting the dense model before sharding, speeds up the distillation or retraining process. The affinity is measured with a first-order approximation of the damage that losing a head's contribution does to each neuron's output, and the co-located affinity is maximised with a coordinate-ascent optimiser that alternates an exact balanced assignment of neurons with an exhaustive search over the KV head partitions. The whole procedure takes under two minutes on one GPU for \qwen{} and \danube. On these models, at $\delta=1$, the affinity-optimised layouts reach any distillation target in about half to two thirds of the steps needed by the naive contiguous layouts, over the whole 10k-step range we tested, and every optimised seed beats every contiguous seed and all but one of the sixteen random layouts. We also show that the co-located affinity score at initialisation predicts the KL to the base model after training, across seventeen layouts ranging from anti-optimised to optimised (Pearson $-0.81$ and $-0.89$). Ablations show that the gain comes mostly from placing neurons (more than heads), and that zero-shot damage (as opposed to zero-shot affinity) is a poor predictor of the trained quality.
\end{abstract}

\section{Introduction}
\label{sec:intro}

Batch-size-one decoding of a Transformer is bound by weight streaming, and can therefore be sped up with tensor parallelism (TP), which divides the amount of weights to be streamed on each device by the total number of devices \citep{shoeybi2019megatron}. Communication overhead soon becomes an issue at low batch sizes, as the computations themselves become fast and shorter than the communication. At low batch sizes again, the communication speed is limited by incompressible latency more than by bandwidth itself. As a consequence, the all-reduce becomes a large share of the per-token latency: removing it improves the decode throughput of Llama-3.1-8B on eight H200s by 43\% \citep{taneja2026sifar}. Various works focus on changing the architecture to hide communication behind compute:
\begin{itemize}
\item parallel attention and MLP blocks halve the number of synchronisations, at the cost of also halving the critical-path depth \citep{wang2021gptj,chowdhery2023palm};
\item Ladder Residual feeds each module the previous module's residual, so that compute happens while the all-reduce finishes \citep{zhang2025ladder};
\item Kraken and Parallel Track Transformers run several thinner Transformers side by side with periodic syncs, at the cost of lowering the effective width of each layer \citep{prabhakar2024kraken,wang2026pt};
\item Sync-Point Drop skips the attention all-reduce in insensitive blocks, and therefore sometimes all-reduces only after the FFN \citep{kim2025spd};
\item CAAT-Net synchronises only a subset of channels \citep{lamprecht2025caat}.
\end{itemize}

Delayed Tensor Parallelism (\dtp) \citep{kog2026dtp} is the most recent addition, and our paper builds on it. In \dtp, every device adds its own partial output to its residual stream (and broadcasts it) immediately, but only gathers (receives) the other devices' partials $\delta$ modules later. At $\delta=1$, and from the second module up to the penultimate one, the input of each module is the full residual stream minus the other devices' contribution to the immediately preceding module. In other words, the input of module $n$ on device $d$ is the residual stream of device $d$ after module $n-1$, plus the contribution of all other devices after module $n-2$.

As \dtp{} changes the function the network computes, a pretrained dense model has to be retrained or distilled after adaptation, like every other architecture mentioned above \citep{kim2025spd,zhang2025ladder}. \dtp{} also breaks the permutation symmetry of neurons inside FFNs and of KV heads inside attention modules, and therefore makes the sharding a non-neutral choice that can be optimised.

Under \dtp, the FFN of layer $n$ on device $d$ only sees the attention output of its co-located set of preceding KV heads. If the neurons whose output depends most on these KV heads indeed sit on the same device, little information is lost. If, on the contrary, the neurons most sensitive to these KV heads' output lie on other devices, most of the information they output will be lost for layer $n$.

Since, for the original dense model, the heads of an attention layer and the neurons of an FFN are permutation-symmetric \citep{entezari2022permutation,ainsworth2023git,tran2026functional}, we are free to choose any balanced assignment of heads and neurons to devices. The permutation only changes the \dtp{} function, while leaving the dense model unchanged to within floating-point error.

\paragraph{Contributions}
\begin{enumerate}
\item We formalise the choice of sharding under \dtp{} as a layout optimisation problem over permutations of key-value (KV) groups and FFN neurons, and give a calibration-based affinity score between heads and neurons (a first-order estimate of the damage to the residual stream when a neuron loses one head's contribution) and between neurons and the next layer's KV groups (\S\ref{sec:method}).
\item We give a fast optimiser for the resulting chain of assignment problems: an exact balanced linear assignment for the neurons of a layer given its neighbouring head partitions, and an exhaustive search over the 2520 labelled balanced partitions of eight KV groups into four devices given the neighbouring neuron assignments, alternated until convergence. Affinity pass plus optimisation take under two minutes on one GPU.
\item On two pretrained models sharded over four devices at $\delta=1$, distilled from the dense model with an identical recipe for every layout, we show that (a) every optimised seed beats every contiguous seed and every random layout on \qwen{} (seven of eight on \danube), and the optimised layout reaches the contiguous run's 1000-step quality in 347 steps on \qwen{} and 565 on \danube{}; (b) over 10k steps the optimised run reaches each contiguous KL value in about half to two thirds of the steps, with no sign of the gap closing on \qwen{}; (c) the co-located affinity of a layout predicts its trained KL across seventeen layouts; (d) the gain lives in the neuron assignment, the head assignment adds nothing measurable, and the zero-shot damage of a layout is uninformative about its trained quality (\S\ref{sec:results}).
\end{enumerate}

\section{Background: Delayed Tensor Parallelism at $\delta=1$}
\label{sec:background}

We consider a pre-norm decoder with $N$ layers, each an attention module followed by an FFN module, indexed jointly by $n=0,\dots,2N-1$ (attention of layer $i$ is module $2i$, its FFN module $2i+1$). Sharded over $L$ devices in the Megatron way \citep{shoeybi2019megatron}, attention is split by heads (by KV groups under grouped-query attention \citep{ainslie2023gqa}, each group carrying its query heads) and the FFN by intermediate neurons, so that module $n$ on device $l$ produces a partial output $o^{n}_l$ and the vanilla module output is $\sum_l o^{n}_l$. Standard TP all-reduces after every module; each device then continues from the same residual stream.

\begin{figure}[t]
\centering
\begin{tikzpicture}[x=1cm,y=1cm,
  dev/.style={draw,rounded corners=2pt,fill=blue!6,text width=3.0cm,align=center,minimum height=0.55cm,font=\small,inner sep=2pt},
  miss/.style={draw,rounded corners=2pt,fill=red!8,dashed,text width=3.0cm,align=center,minimum height=0.55cm,font=\small,inner sep=2pt},
  wide/.style={draw,rounded corners=2pt,fill=blue!6,text width=10.4cm,align=center,minimum height=0.55cm,font=\small,inner sep=2pt},
  lbl/.style={font=\small\itshape,text=black!65},
  arr/.style={-{Latex[length=2mm]},thick}]
  \node[wide,fill=gray!12] (S) at (0,2.4) {$S^{(2i-1)}$: every device's partials up to the FFN of layer $i-1$};
  \node[dev]  (a0) at (-3.6,1.2) {heads on device $l$};
  \node[miss] (a1) at (0,1.2)    {heads on device $j\ne l$};
  \node[miss] (a2) at (3.6,1.2)  {heads on device $j'\ne l$};
  \node[lbl,anchor=east] at (-5.5,1.2) {attention $i$};
  \node[wide] (F) at (0,0) {FFN of layer $i$, the neurons on device $l$};
  \node[lbl,anchor=east] at (-5.5,0) {FFN $i$};
  \draw[arr] (S.south -| a0) -- (a0.north);
  \draw[arr] (S.south -| a1) -- (a1.north);
  \draw[arr] (S.south -| a2) -- (a2.north);
  \draw[arr] (a0.south) -- (F.north -| a0);
  \draw[arr,red!70,dashed] (a1.south) -- (F.north -| a1);
  \draw[arr,red!70,dashed] (a2.south) -- (F.north -| a2);
  \node[dev]  (b0) at (-3.6,-1.2) {KV groups on device $l$};
  \node[miss] (b1) at (0,-1.2)    {KV groups on device $j\ne l$};
  \node[miss] (b2) at (3.6,-1.2)  {KV groups on device $j'\ne l$};
  \node[lbl,anchor=east] at (-5.5,-1.2) {attention $i+1$};
  \draw[arr] (F.south -| b0) -- (b0.north);
  \draw[arr,red!70,dashed] (F.south -| b1) -- (b1.north);
  \draw[arr,red!70,dashed] (F.south -| b2) -- (b2.north);
  \node[font=\scriptsize,text=red!70] at (0,-2.0) {dashed: the other devices' partial of the preceding module, which lands one module later};
\end{tikzpicture}
\caption{What device $l$ sees under \dtp{} at $\delta=1$ (equation~\ref{eq:delta1}). Solid: available when the module runs. Dashed red: not yet available. The two edges the layout can control are heads$(i)\to$neurons$(i)$ (the \emph{up} edge) and neurons$(i)\to$KV groups$(i+1)$ (the \emph{down} edge).}
\label{fig:schematic}
\end{figure}
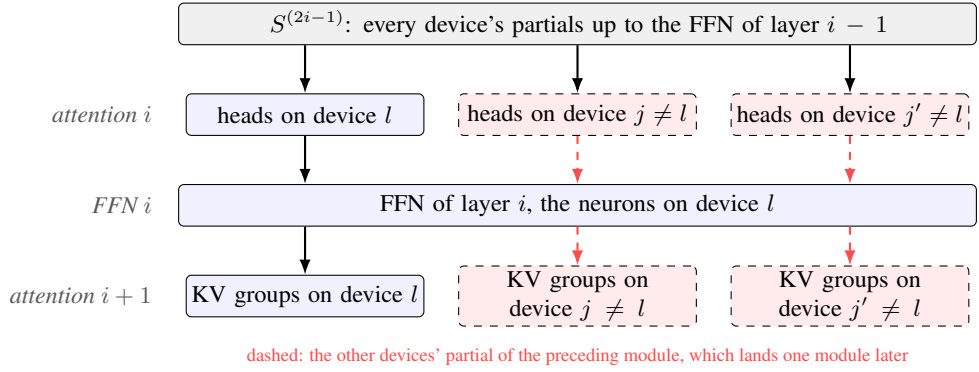

\dtp{} gives each device its own residual stream $X_l$ and delays the aggregation by $\delta$ modules \citep{kog2026dtp}:
\begin{align}
n<\delta:\qquad & X_l \leftarrow X_l + \sqrt{L}\,o^{n}_l, \label{eq:stage1}\\
\delta\le n<2N-\delta:\qquad & X_l \leftarrow X_l + o^{n}_l + \textstyle\sum_{j\ne l} o^{\,n-\delta}_j, \label{eq:stage2}\\
n\ge 2N-\delta:\qquad & X_l \leftarrow X_l + \sqrt{L}\,o^{n}_l + \textstyle\sum_{j\ne l} o^{\,n-\delta}_j, \label{eq:stage3}
\end{align}
where the $\sqrt{L}$ factor in the first and last $\delta$ modules mimics the magnitude of a full all-reduce. At the end the final norm is applied per device and the $L$ outputs are averaged. The broadcast of $o^n_l$ has $\delta$ modules of weight streaming to arrive, which is what removes the exposed communication. This paper studies $\delta=1$, the smallest delay and the one where the structure is cleanest. Unrolling \eqref{eq:stage2} at $\delta=1$ shows that before module $n+1$ device $l$ holds
\begin{equation}
X^{(n)}_l \;=\; S^{(n-1)} + o^{n}_l, \qquad S^{(m)} \;=\; X^{(0)} + \textstyle\sum_{k\le m}\sum_j o^{k}_j ,
\label{eq:delta1}
\end{equation}
that is, the sum of \emph{every} device's partials up to module $n-1$, plus only its \emph{own} partial of module $n$ (Figure~\ref{fig:schematic}). Two things follow. The FFN of layer $i$ on device $l$ sees the attention of layer $i$ only through the KV groups placed on $l$; the attention of layer $i+1$ on device $l$ sees the FFN of layer $i$ only through the neurons placed on $l$. Everything older is complete. The partials themselves are computed from diverged streams, so the model is not the dense model and must be trained; but what a device misses is determined entirely by which heads and neurons share the device.

\section{Method: affinity-aware layouts}
\label{sec:method}

\paragraph{The layout as a free variable}
For layer $i$ let $h_i:\{1..G\}\to\{1..L\}$ assign its $G$ KV groups to devices and $\nu_i:\{1..I\}\to\{1..L\}$ assign its $I$ FFN neurons, both balanced ($G/L$ groups and $I/L$ neurons per device). Any such pair is realised by permuting the rows of the query, key and value projections (by group, keeping each group's query heads with its KV heads) and the columns of the output projection identically, and the rows of the gate and up projections and the columns of the down projection identically, then slicing contiguous shards. The dense function is unchanged: we verify a maximum logit difference of $10^{-4}$ in fp32. The rotary embedding, the per-head norms of Qwen3 and the SwiGLU nonlinearity \citep{shazeer2020glu} are all applied per head or per neuron, so they commute with the permutation.

\paragraph{Affinity scores}
Per layer we need two matrices: $S^{\uparrow}_i\in\mathbb{R}^{G\times I}$, how much neuron $k$ of layer $i$ depends on KV group $g$ of layer $i$, and $S^{\downarrow}_i\in\mathbb{R}^{I\times G}$, how much KV group $g$ of layer $i+1$ depends on neuron $k$ of layer $i$. Both are estimated from the dense model on calibration text (64 sequences of 1024 tokens of FineWeb-Edu \citep{penedo2024fineweb}, one forward pass, no gradients). We write $x$ for the vanilla residual entering the FFN's pre-norm, $r=\mathrm{rms}(x)$, $\gamma$ the norm gain, $a_h$ head $h$'s attention output and $W_o^{(h)}$ its slice of the output projection, so that the normalised FFN input decomposes additively over heads as $\sum_h p_h + (\text{rest})$ with $p_h = \gamma\odot(a_h W_o^{(h)})/r$ \citep{elhage2021mathematical}. Neuron $k$'s pre-activations therefore decompose as $g_k=\sum_h p_h\!\cdot\! w^{g}_k+\dots$ and $u_k=\sum_h p_h\!\cdot\! w^{u}_k+\dots$, and the first-order change of its output $\mathrm{silu}(g_k)\,u_k$ when head $h$'s piece is removed is
\begin{equation}
\Delta_{hk} \;=\; \mathrm{silu}'(g_k)\,u_k\,(p_h\!\cdot\! w^{g}_k) \;+\; \mathrm{silu}(g_k)\,(p_h\!\cdot\! w^{u}_k).
\end{equation}
Our main score, \fo{} (first order), is the mean squared damage this does to the residual stream, $S^{\uparrow,\fo}_{hk}=\mathbb{E}_t[\Delta_{hk}^2]\,\|w^{d}_k\|^2$ with $w^d_k$ the down-projection column, summed over the query heads of each KV group. The simpler \addsc{} score is the mean squared additive contribution to the pre-activations, $\mathbb{E}_t[(p_h\!\cdot\! w^{g}_k)^2+(p_h\!\cdot\! w^{u}_k)^2]$; it ignores the operating point of the nonlinearity and the size of the neuron's write. For the down edge, neuron $k$ writes $\alpha_k w^d_k$ into the residual ($\alpha_k$ its activation), which enters the next layer's q/k/v projections of group $g$ through that layer's pre-norm (gain $\gamma'$, rms $r'$) as $b_{gk}=W^{qkv}_g(\gamma'\odot w^d_k)$, so $S^{\downarrow}_{kg}=\mathbb{E}_t[\alpha_k^2/r'^2]\,\|b_{gk}\|^2$. We also computed exact leave-one-out ablation scores of the pre-activations, which differ from \addsc{} only by including the change of the norm's rms when a head is removed; their Spearman correlation with \addsc{} is above 0.98 on every layer but one (0.83 on the last layer of \danube), so the rms change is negligible and both \addsc{} and \fo{} ignore it.

\paragraph{Objective}
Each matrix is normalised to unit mass so that every edge counts equally, and the layout score is the total co-located affinity,
\begin{equation}
J(h,\nu)\;=\;\sum_{i=0}^{N-1}\ \sum_{g,k:\;h_i(g)=\nu_i(k)} \hat S^{\uparrow}_i[g,k]\;+\;\sum_{i=0}^{N-2}\ \sum_{k,g:\;\nu_i(k)=h_{i+1}(g)} \hat S^{\downarrow}_i[k,g],
\label{eq:objective}
\end{equation}
which reads as ``number of edges times mean co-located fraction'': it ranges over $[0,\,2N-1]$ and a random balanced layout scores $(2N-1)/L$ in expectation. $J$ is a proxy, not the training loss; \S\ref{sec:score} tests how well it predicts the trained outcome.

\paragraph{Optimiser}
$J$ couples the layers in a chain (Figure~\ref{fig:schematic}), and each block of the chain is easy given its neighbours. Given $h_i$ and $h_{i+1}$, the gain of putting neuron $k$ on device $l$ is
\[
M_{kl}\;=\;\sum_{g:\,h_i(g)=l}\hat S^{\uparrow}_i[g,k]\;+\;\sum_{g:\,h_{i+1}(g)=l}\hat S^{\downarrow}_i[k,g],
\]
and the best balanced $\nu_i$ is a linear assignment problem on $I/L$ replicated slots per device \citep{kuhn1955hungarian,crouse2016lsa}. We solve it on the GPU by dual ascent on $L$ device offsets, a fix-up to meet the capacities exactly, and pairwise swaps until no swap improves; on the layers we checked it matches the Hungarian solution to $10^{-6}$. Given $\nu_{i-1}$ and $\nu_i$, the best $h_i$ is found by scoring all $G!/((G/L)!)^L$ labelled balanced partitions of the KV groups (2520 for $G=8$, $L=4$). Starting from the contiguous head assignment we alternate the two steps over all layers until $J$ stops increasing, which takes two to four sweeps. Restricting which layers move, freezing heads or neurons, or minimising $J$ instead gives the ablation layouts of \S\ref{sec:ablations}.

\paragraph{Cost}
The affinity pass takes 49 s (\danube) and 51 s (\qwen) on one RTX~5090 for 65,536 tokens; the optimisation 38 to 47 s including scoring eight random layouts. One 1000-step distillation run takes 22 and 32 min on the same GPU, so the method costs about 7\% of the shortest run in this study.

\section{Experimental setup}
\label{sec:setup}

\paragraph{Models and sharding}
\qwen{} \citep{qwen3} (28 layers, 16 query and 8 KV heads, head dim 128, FFN width 3072) and \danube{} \citep{danube3} (16 layers, 16 query and 8 KV heads, FFN 4096), both pre-norm SwiGLU decoders with the Llama module layout. $L=4$ virtual devices are simulated on one GPU (the sharding arithmetic is exact and tested; the speed-up is not measured here), $\delta=1$, $\sqrt{L}$ own-scaling in the first and last module. Every layout is balanced: two KV groups and $I/4$ neurons per device. \emph{Contiguous} is the standard layout that slices heads and neurons in storage order; \emph{random} layouts are uniform balanced permutations; \emph{optimised} maximises \eqref{eq:objective} with the \fo{} score.

\paragraph{Training}
All layouts are trained with the same recipe: distillation from the frozen dense model, loss $\KL(\text{teacher}\,\|\,\text{student})+0.1\,\mathrm{CE}$ per token \citep{hinton2015distilling}, AdamW \citep{loshchilov2019adamw} with $\beta=(0.9,0.95)$, no weight decay, learning rate $5\times10^{-5}$ with 100 warm-up steps and cosine decay to $5\times10^{-6}$, gradient clipping at 1, embeddings frozen, fp32 master weights with bf16 autocast. Each step is 16 sequences of 1024 tokens of FineWeb-Edu (16k tokens); the same pre-tokenised file is used for every run and the training seed only changes the block order. Short runs are 1000 steps (16M tokens), long runs 10,000 steps (164M tokens).

\paragraph{Evaluation}
Every 100 steps (250 for the long runs) we measure, on 16 blocks of 1024 tokens of the WikiText-2 test set \citep{merity2017pointer}, the perplexity of the \dtp{} model and its per-token $\KL$ to the dense model in nats, which we report as the primary metric because it measures exactly what distillation minimises and does not saturate. The dense models have perplexity 12.79 (\qwen) and 10.70 (\danube) on these blocks. Sixteen blocks are enough for the curves and the correlations but not for differences below about 0.01 KL, and we say so where it matters.

\paragraph{Design}
Per model: three training seeds each for the optimised and contiguous layouts; eight random layouts with one seed each; seven further single-seed layouts that spread the score axis (anti-optimised, i.e. $J$ minimised; heads only; neurons only; \addsc{} score; and 4, 8 or 12 evenly spaced layers optimised with the rest contiguous); and one 10k-step run each for optimised and contiguous. That is 23 training runs per model, 46 in all, plus two affinity passes and sixteen layout optimisations, about 19 hours of wall clock on two RTX~5090s.

\section{Results}
\label{sec:results}

\subsection{The optimised layout trains faster and better, on every seed}
\label{sec:seeds}

\begin{table}[t]
\centering
\footnotesize
\setlength{\tabcolsep}{4pt}
\caption{KL to the dense model (nats per token, mean $\pm$ sample sd over seeds) and perplexity on the 16 WikiText-2 blocks after 1000 distillation steps, and the step at which the mean optimised curve first reaches the contiguous arm's final KL. Random is eight layouts with one seed each.}
\label{tab:main}
\begin{tabular}{llccccc}
\toprule
Model & Layout & $n$ & KL@500 & KL@1000 & ppl@1000 & step matching contig.@1000 \\
\midrule
\multirow{3}{*}{\qwen} & optimised (\fo) & 3 & $0.361\pm0.009$ & $\mathbf{0.303\pm0.003}$ & $\mathbf{17.01}$ & 347 ($2.9\times$) \\
 & contiguous & 3 & $0.500\pm0.055$ & $0.415\pm0.033$ & 19.20 & 1000 \\
 & random & 8 & $0.449\pm0.044$ & $0.374\pm0.015$ & 18.31 & --- \\
\midrule
\multirow{3}{*}{\danube} & optimised (\fo) & 3 & $0.763\pm0.056$ & $\mathbf{0.667\pm0.043}$ & $\mathbf{16.05}$ & 565 ($1.8\times$) \\
 & contiguous & 3 & $0.832\pm0.010$ & $0.738\pm0.009$ & 17.45 & 1000 \\
 & random & 8 & $0.844\pm0.023$ & $0.737\pm0.018$ & 17.27 & --- \\
\bottomrule
\end{tabular}
\end{table}

\begin{figure}[t]
\centering
\includegraphics[width=0.94\linewidth]{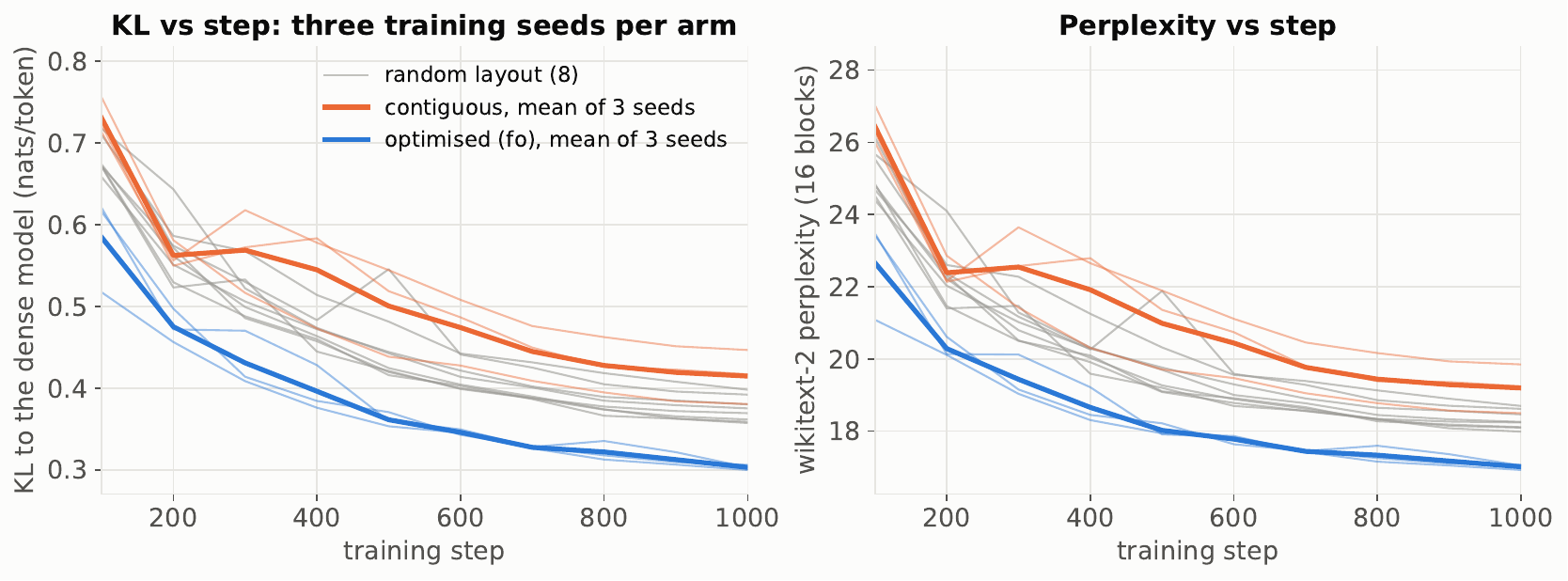}
\caption{\qwen, $L=4$, $\delta=1$: KL to the dense model (left) and perplexity (right) versus distillation step. Bold: mean of three training seeds; faint: individual seeds; grey: the eight random layouts. The \danube{} version is Figure~\ref{fig:danube_seeds} in the appendix.}
\label{fig:seeds}
\end{figure}

Table~\ref{tab:main} and Figure~\ref{fig:seeds} give the headline. On \qwen{} the optimised layout finishes at 0.303 KL against 0.415 for contiguous and 0.374 for random, a 27\% reduction against contiguous and 19\% against random; every optimised seed is below every one of the eleven baseline runs, with a 0.05 KL clearance to the nearest random layout (Welch $t$-test against contiguous $p=0.027$ with $n=3$ per arm; against the eight random layouts $p<0.001$). On \danube{} the margin is smaller, 0.667 against 0.738 (10\%), but again all three optimised seeds are below all three contiguous seeds and below seven of the eight random layouts (Mann-Whitney against random $p=0.024$; Welch against contiguous $p=0.10$ at $n=3$). The mean optimised curve reaches the contiguous arm's final KL at step 347 on \qwen{} and 565 on \danube.

Two remarks on the noise. Contiguous is not a privileged layout: on \danube{} it is indistinguishable from random (0.738 versus 0.737) and on \qwen{} it is, if anything, worse (0.415 versus 0.374), so the standard sharding is just one draw from the random distribution as far as \dtp{} is concerned. And the larger seed spreads (contiguous on \qwen, optimised on \danube) are an evaluation-set property rather than a training one: in both cases the seeds' training losses agree to three decimals, and the same seed under the 10k-step schedule reads within the tight cluster at step 1000. Sixteen blocks are noisy at the 0.03 level; the ordering holds either way.

\subsection{The co-located affinity predicts the trained outcome}
\label{sec:score}

\begin{figure}[t]
\centering
\includegraphics[width=\linewidth]{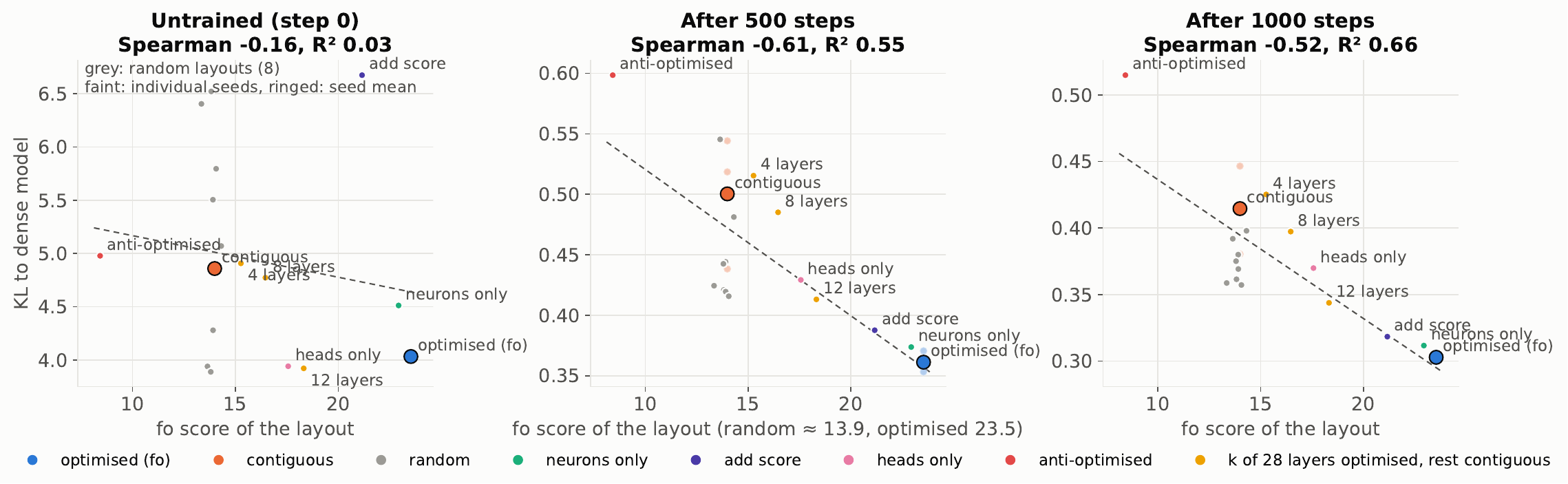}\\[2pt]
\includegraphics[width=\linewidth]{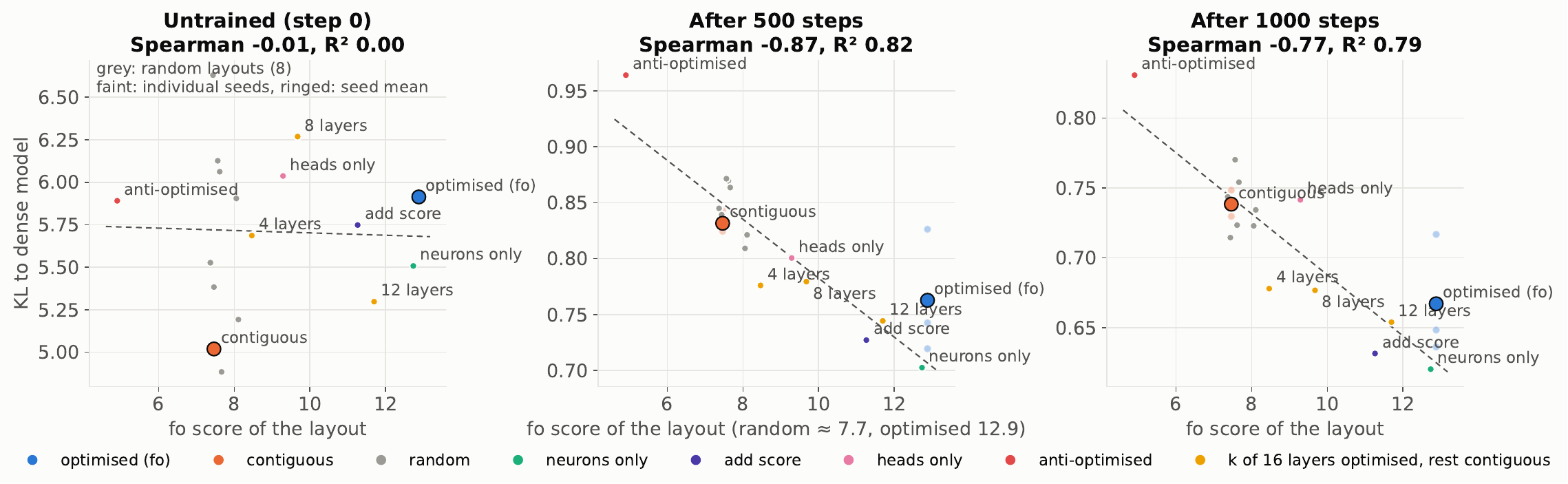}
\caption{KL to the dense model against the co-located \fo{} score $J$ of the layout (equation~\ref{eq:objective}), untrained (left), after 500 (middle) and 1000 steps (right); \qwen{} top, \danube{} bottom. One point per layout, seeds averaged where there are several (ringed), individual seeds faint; dashed line: least-squares fit over the seventeen layouts. Untrained, the score says nothing; after training it explains most of the variance.}
\label{fig:score}
\end{figure}

The objective \eqref{eq:objective} is a proxy, and the eight random layouts alone cannot test it: their scores sit within one unit of each other and their trained KLs differ by evaluation noise. The seven spread layouts (anti-optimised, layer subsets, heads only, \addsc, neurons only) cover the score axis from 8.4 to 23.5 on \qwen{} (random $\approx13.9$) and 4.9 to 12.9 on \danube{} (random $\approx7.7$). Figure~\ref{fig:score} plots the trained KL of all seventeen layouts against their score. After 1000 steps the Pearson correlation is $-0.81$ on \qwen{} ($R^2=0.66$, $p<0.001$) and $-0.89$ on \danube{} ($R^2=0.79$); at 500 steps $-0.74$ and $-0.91$. The Spearman rank correlation is lower on \qwen{} ($-0.52$) because half the points are random layouts whose ranks are noise; over the nine non-random layouts it is $-0.98$ at both 500 and 1000 steps. The negative control matters: the anti-optimised layout, which minimises $J$, is the worst run in either study, 0.14 KL above random on \qwen{} and 0.09 on \danube, so the direction of the objective, not merely ``structure versus none'', drives the outcome. Untrained (left panels) the correlation is zero on both models; we return to this in \S\ref{sec:untrained}.

\subsection{The gap does not close within 10k steps}
\label{sec:long}

\begin{figure}[t]
\centering
\includegraphics[width=0.94\linewidth]{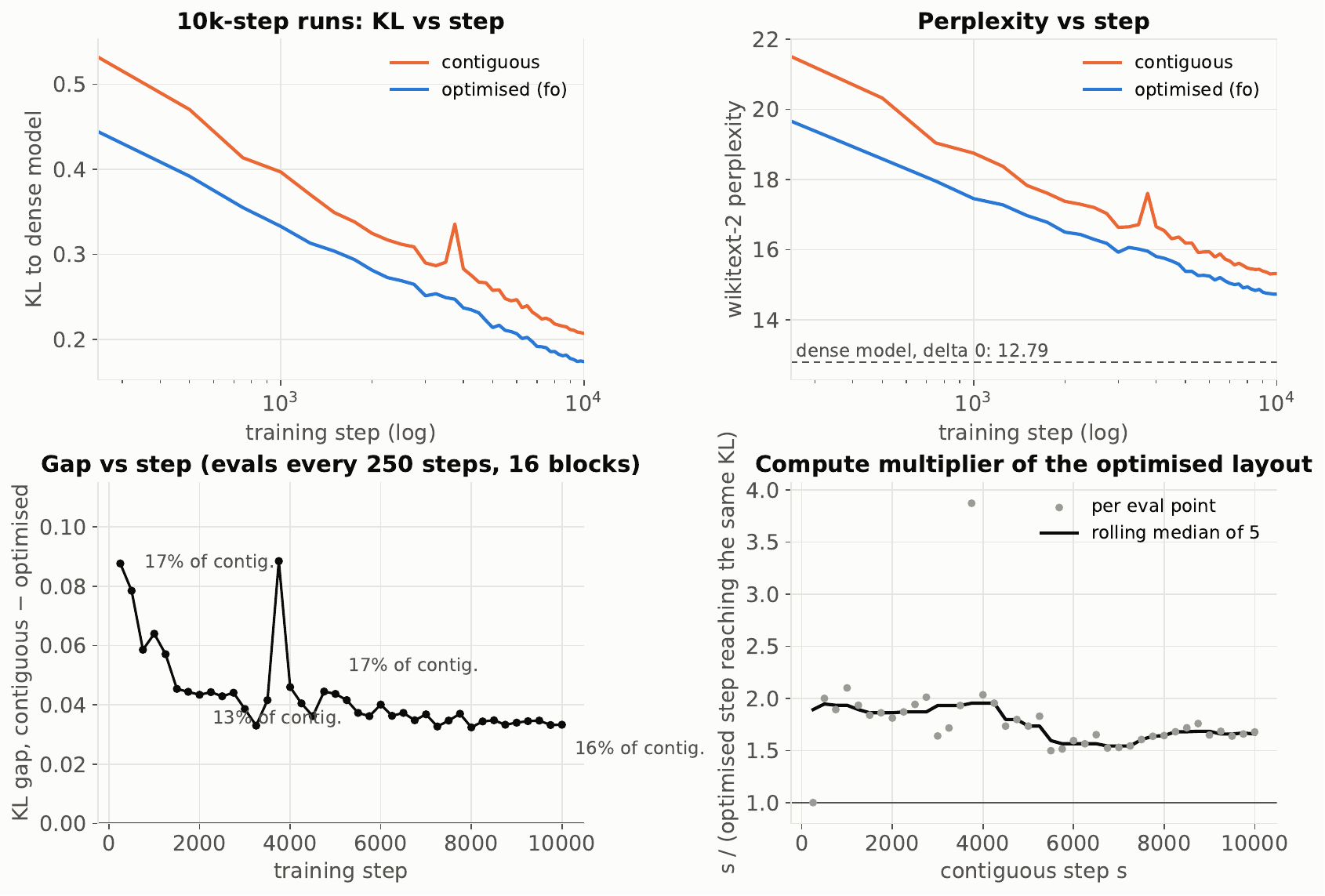}
\caption{\qwen, 10,000 distillation steps (164M tokens), one seed, evaluation every 250 steps. Top: KL and perplexity on a log step axis. Bottom left: the KL gap contiguous minus optimised with its size relative to the contiguous KL. Bottom right: the compute multiplier, the contiguous step $s$ divided by the interpolated step at which the optimised run first reached the same KL (grey per evaluation, black rolling median of five). \danube: Figure~\ref{fig:danube_gap}.}
\label{fig:gap}
\end{figure}

A better initialisation could be a warm start that training washes out. Figure~\ref{fig:gap} shows the two 10k-step runs on \qwen. The optimised run is ahead at all 40 evaluation points. The absolute gap falls with the KL itself, from 0.088 at step 250 to 0.033 at step 10,000, but as a fraction of the contiguous KL it stays between 13\% and 17\% at every tabulated step (Table~\ref{tab:long}), and the last eight evaluations all read between 15.5\% and 16.6\%. Read as compute, the optimised run reaches each contiguous KL value 1.6 to 2.1$\times$ earlier at the tabulated steps of Table~\ref{tab:long}, and 1.5 to 2.1$\times$ earlier at every evaluation from step 1000 on, apart from a single-evaluation spike of the contiguous run at step 3750 (1.68$\times$ at step 10,000). On \danube{} (appendix, Figure~\ref{fig:danube_gap}) the relative gap does shrink, from 16\% at step 500 to 4.6\% at 10,000, but the optimised run is again ahead at all 40 points. Its multiplier is noisier: 1.6 to 2.4$\times$ at the tabulated steps, 1.3 to 3.2$\times$ per evaluation, with a rolling median that falls from 2.7$\times$ early on to 1.45$\times$ around step 7000 and ends at 1.6$\times$ (1.64$\times$ at step 10,000). Both curves are still falling at the floor learning rate, so neither number is an asymptote. The claim that both models support is that the optimised layout reaches any distillation target in about half to two thirds of the steps over the whole range we ran; whether a residual advantage survives at convergence is model-dependent and one seed per model cannot settle it.

\begin{table}[t]
\centering
\small
\caption{The two 10k-step runs on \qwen{} (one seed). The multiplier is the contiguous step $s$ over the optimised step reaching the same KL.}
\label{tab:long}
\begin{tabular}{rcccccc}
\toprule
step & optimised KL & contiguous KL & gap & gap / contiguous & opt.\ ppl / contig.\ ppl & multiplier \\
\midrule
   500 & 0.392 & 0.470 & 0.079 & 16.7\% & 18.59 / 20.32 & $\ge2.0$ \\
  1000 & 0.333 & 0.397 & 0.064 & 16.1\% & 17.46 / 18.75 & 2.10 \\
  2000 & 0.281 & 0.325 & 0.043 & 13.4\% & 16.50 / 17.38 & 1.81 \\
  4000 & 0.237 & 0.283 & 0.046 & 16.3\% & 15.81 / 16.66 & 2.03 \\
  6000 & 0.207 & 0.247 & 0.040 & 16.3\% & 15.25 / 15.94 & 1.59 \\
  8000 & 0.186 & 0.218 & 0.032 & 14.9\% & 14.94 / 15.48 & 1.64 \\
 10000 & 0.174 & 0.207 & 0.033 & 16.1\% & 14.73 / 15.32 & 1.68 \\
\bottomrule
\end{tabular}
\end{table}

\subsection{Ablations: neurons carry the gain, heads add nothing measurable}
\label{sec:ablations}

\begin{table}[t]
\centering
\small
\caption{Ablation layouts (single seed unless noted), ordered by \qwen{} score. Score $J$ is the co-located \fo{} affinity (equation~\ref{eq:objective}); its random expectation is 13.75 on \qwen{} (55 edges) and 7.75 on \danube{} (31 edges). KL after 1000 steps. Differences below 0.01 are within evaluation noise.}
\label{tab:ablation}
\begin{tabular}{llcccc}
\toprule
 & & \multicolumn{2}{c}{\qwen} & \multicolumn{2}{c}{\danube} \\
\cmidrule(lr){3-4}\cmidrule(lr){5-6}
Layout & What moves & $J$ & KL@1000 & $J$ & KL@1000 \\
\midrule
anti-optimised & everything, $J$ minimised & 8.42 & 0.515 & 4.91 & 0.831 \\
contiguous (3 seeds) & nothing & 13.99 & 0.415 & 7.46 & 0.738 \\
random (8 layouts) & everything, random & $\approx13.9$ & 0.374 & $\approx7.7$ & 0.737 \\
4 layers & evenly spaced layers, incl.\ layer 0 & 15.27 & 0.425 & 8.46 & 0.678 \\
8 layers & & 16.46 & 0.397 & 9.67 & 0.677 \\
heads only & KV groups; neurons contiguous & 17.57 & 0.370 & 9.29 & 0.742 \\
12 layers & & 18.32 & 0.344 & 11.69 & 0.654 \\
\addsc{} score & everything, dot-product objective & 21.16 & 0.318 & 11.26 & 0.632 \\
neurons only & neurons; KV groups contiguous & 22.94 & 0.312 & 12.73 & 0.620 \\
optimised, \fo{} (3 seeds) & everything & 23.54 & 0.303 & 12.88 & 0.667 \\
\bottomrule
\end{tabular}
\end{table}

\begin{figure}[t]
\centering
\includegraphics[width=0.94\linewidth]{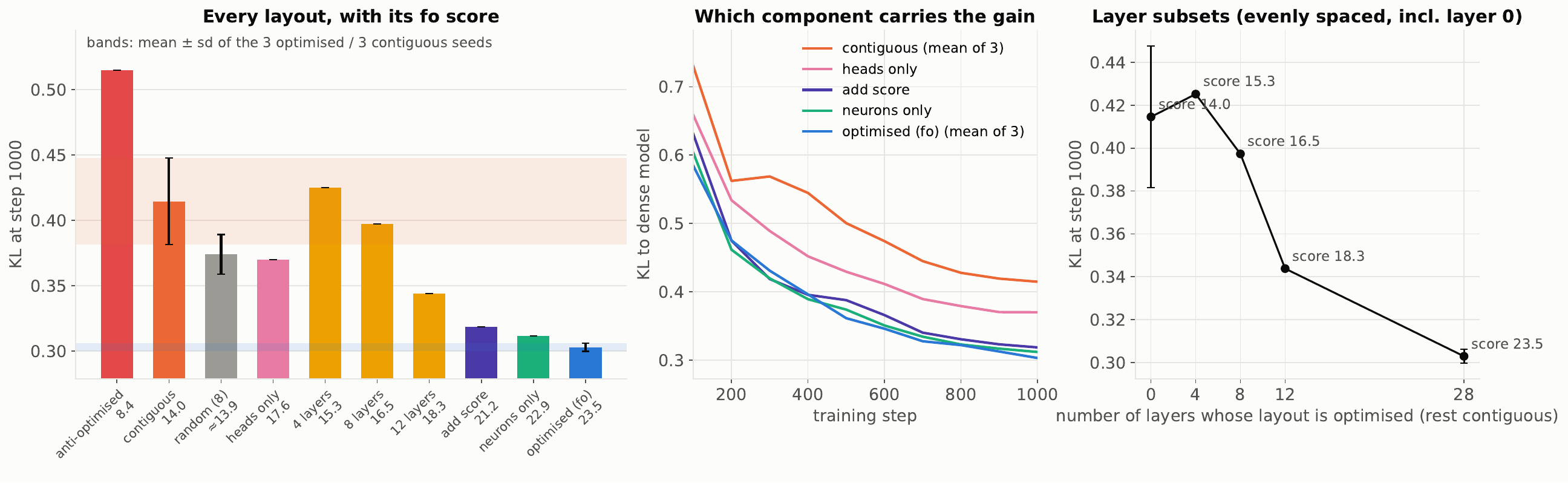}
\caption{\qwen{} ablations. Left: KL at step 1000 for every layout, ordered by score (in the tick labels); shaded bands are mean $\pm$ sd of the three optimised (blue) and three contiguous (orange) seeds. Middle: KL curves of the component ablations against the two seed means. Right: KL at 1000 steps against the number of layers optimised (evenly spaced, always including layer 0).}
\label{fig:ablations}
\end{figure}

Table~\ref{tab:ablation} and Figure~\ref{fig:ablations} decompose the gain.

\emph{Neurons, not heads.} Moving only the neurons (KV groups contiguous) captures essentially the whole effect on both models: 0.312 against 0.303 on \qwen{} and 0.620 against 0.667 on \danube, where it is in fact the best 1000-step number in the study. Moving only the KV groups (neurons contiguous) reaches a score half-way from random to optimised on \qwen{} (17.6) but trains to the random mean to three decimals (0.370), and on \danube{} it equals contiguous (0.742). This is consistent with the sizes of the two search spaces: the head step chooses among 2520 partitions of 8 groups, the neuron step among balanced assignments of 3072 or 4096 free variables per layer. The head step can be presented as an optional refinement; in practice one can skip it and keep the standard head sharding.

\emph{First-order versus dot-product affinity.} The \addsc{} layout, rescored in \fo{} units, reaches 87 to 90\% of the optimised score and trains within 0.015 KL of it on \qwen{} and within the seed band on \danube. The first-order score is the principled one and finds a layout with a 4-point higher co-located fraction, but this study cannot separate the two objectives at the 1000-step scale.

\emph{How many layers.} On \danube, optimising 4 of 16 evenly spaced layers already gives 85\% of the gain, 12 layers all of it; on \qwen{} the curve is close to linear in the number of layers and 4 of 28 gives nothing. The per-layer co-located fractions (appendix, Figures~\ref{fig:qwen_layers} and~\ref{fig:danube_layers}) explain the difference: on \danube{} the optimiser finds two layers (1 and 15) with fractions above 0.6 and the rest near random, so a subset that includes the right layer captures most of the score; on \qwen{} the optimised fraction is between 0.41 and 0.71 on every layer, with one layer at 1.00, and the score, and the gain, accumulate layer by layer. In both models the layer-subset points sit above the fit line of Figure~\ref{fig:score}: the score weights every edge equally and training does not, with the early layers counting for more than their share. The subset design confounds the count of layers with their identity; the clean test (high- versus low-fraction subsets at equal count) is three runs and is left for future work.

\subsection{Zero-shot damage is uninformative}
\label{sec:untrained}

A natural shortcut would be to pick the layout with the smallest untrained perplexity. It does not work. Untrained, the optimised layout is 5th of 17 on \qwen{} (KL 4.03, perplexity 731; contiguous 4.86, 1666; random 3.9 to 6.5) and 12th of 17 on \danube{} (5.91, 3557; contiguous 5.02, 1506), so on one model the standard layout is the best untrained and the optimised one is below median. The Spearman correlation between a layout's untrained KL and its KL after 1000 steps is $+0.19$ on \qwen{} and $-0.24$ on \danube{} (neither significant), and between the score and the untrained KL it is $-0.16$ and $-0.01$ (Figure~\ref{fig:score}, left panels; appendix Figure~\ref{fig:untrained}). By step 100, the end of warm-up, the optimised run is already ahead on both models and the order never flips again. The zero-shot damage of an untrained \dtp{} model is dominated by a few large, easily repaired disruptions; what the layout controls is how much of the network's fine structure the devices can reconstruct once those are repaired.

\subsection{Where the affinity concentrates}
\label{sec:circuit}

The \fo{} score is not uniformly informative across the network. On \qwen{} the optimiser drives the co-located fraction of layer 2's up edge to 1.00 (the anti-optimised layout drives it to 0.00), while \addsc{} reaches only 0.46 there. Probing the dense model shows why: six FFN neurons of layer 2 write an activation of roughly 7000 into a single residual dimension on delimiter tokens, driven almost entirely by two query heads of one KV group. This is the massive-activation circuit described by \citet{sun2024massive}, which originates in the early-layer FFN on delimiter tokens and feeds the attention sinks of later layers \citep{xiao2024streaming,sun2026spike,yu2024super}. The first-order score sees the circuit because it measures the damage in residual-stream units through the SwiGLU operating point; the pre-activation dot product does not, because the neurons' pre-activations are not unusual, only their gain is. More generally, the affinity matrices are far from uniform: averaged over neurons, the single most influential KV group carries 27 to 64\% of a neuron's up-edge affinity (\qwen, per layer; 27 to 48\% on \danube) against 12.5\% if the eight groups contributed equally, which is the structure that \citet{knittel2026sparse} report as sparse inter-layer dependencies of FFN neurons on attention outputs, and that \citet{neo2024interpreting} describe for next-token neurons driven by specific heads.

\section{Related work}
\label{sec:related}

\paragraph{Architectures that communicate less}
Parallel attention and FFN blocks \citep{wang2021gptj,chowdhery2023palm} halve the all-reduces per layer. Ladder Residual \citep{zhang2025ladder} routes each module's input from the previous module's residual so the all-reduce overlaps with compute, and converts Llama-3.1-8B with 3B tokens of retraining. Kraken \citep{prabhakar2024kraken} and Parallel Track Transformers \citep{wang2026pt} run several thinner Transformers in parallel and fuse them periodically, reducing syncs by up to 16$\times$; CAAT-Net \citep{lamprecht2025caat} all-reduces only a subset of channels. \dtp{} \citep{kog2026dtp} keeps the dense Transformer's width and per-module structure and delays the reduction instead. All of these leave each device with partial information for part of the forward pass, and none chooses which heads and neurons a device holds; our contribution is that choice, and it applies in principle to any of them.

\paragraph{Sync-Point Drop}
The closest prior work is SPD \citep{kim2025spd}, which deletes the attention all-reduce in blocks that tolerate it and, for the most sensitive blocks, re-initialises the sharding before block-to-block distillation. SPD's initialisation \emph{scatters} heads across devices by maximising the distance between the attention patterns of heads that share a device, so that every device holds a functionally diverse set, and then matches each head group with one of the existing MLP partitions by output norm. Ours differs in objective and granularity: we \emph{co-locate} heads with the neurons that depend on them, we place individual neurons rather than whole partitions, and our ablations show the neuron assignment is what carries the gain. SPD reports a 3\% accuracy recovery from its head grouping on LLaMA2-7B; a direct comparison under \dtp{} is future work.

\paragraph{Systems approaches}
Kernel-level overlap \citep{chang2024flux,wang2024domino}, faster all-reduce primitives \citep{taneja2026sifar} and compressed activations \citep{hansenpalmus2024compression,li2024flash} reduce the cost of the collective without changing the model. They compose with architectural changes such as \dtp{} and with the layout choice studied here.

\paragraph{Permutation symmetry and hardware-aware permutations}
That hidden units can be permuted without changing the function underlies weight matching for model merging \citep{entezari2022permutation,ainsworth2023git}; \citet{tran2026functional} characterise the corresponding symmetries of attention heads under rotary embeddings, which is the head-level symmetry we use. \citet{pool2021channel} permute channels so that a network fits the N:M sparsity pattern of the hardware without accuracy loss; our work is the same move for a communication constraint.

\paragraph{Neuron clustering and modularity}
MoEfication \citep{zhang2022moefication}, emergent modularity \citep{zhang2023emergent} and LLaMA-MoE \citep{zhu2024llamamoe} partition FFN neurons into experts by co-activation or weight similarity to enable conditional computation. We partition neurons by their dependence on specific heads, for co-location rather than routing, and every neuron still runs on every token. The key-value-memory view of the FFN \citep{geva2021kv} and the sparse attention-to-neuron dependencies of \citet{knittel2026sparse} are the structure that makes a good partition exist.

\paragraph{Converting dense checkpoints by distillation}
Uptraining a pretrained model into a cheaper architecture with a short distillation is standard practice \citep{ainslie2023gqa,muralidharan2024minitron,zhang2025ladder,kim2025spd}. Our result is about the starting point of that distillation: a better permutation of the same weights roughly halves the number of steps to a target.

\section{Limitations and future work}
\label{sec:limitations}

Everything here is $L=4$ and $\delta=1$ on two sub-billion-parameter models, and the devices are simulated on one GPU, so we make no claim about wall-clock speed. At larger $\delta$ a module misses the other devices' partials of the last $\delta$ modules and the affinity graph gains edges; the objective and the optimiser extend directly, but we have not run it. The seed studies have $n=3$ per arm and the long runs one seed, evaluated on 16 blocks of WikiText-2; the direction of every comparison is consistent, but differences below 0.01 KL should be read as ties, and the final numbers deserve the full test set and downstream tasks. The layer-subset ablation confounds how many layers move with which ones. Finally, the affinity is measured on the dense model and used once; re-estimating it on the partially trained \dtp{} model, or making the layout part of training, are natural extensions. A shared-expert variant that replicates 10\% of each layer's neurons on every device gave no further gain in a preliminary run (appendix).

\section{Conclusion}

Delayed Tensor Parallelism turns the sharding of a Transformer from a systems detail into a modelling choice: a device can only see the previous module through the heads and neurons it holds. Because heads and neurons are permutation-symmetric, the choice is free. A first-order affinity from one calibration pass and a two-minute assignment optimiser produce a layout that reaches any distillation target in about half the steps of the standard layout, on both models we tried, and the objective it maximises predicts the trained outcome across seventeen layouts. The gain comes from placing FFN neurons next to the heads they depend on. We expect the same principle to apply to every architecture that trades synchronisation for partial information.

\subsubsection*{Acknowledgements}
We thank the Kog team for the \dtp{} design and for discussions. A large language model was used to draft the text and figures of this paper from the author's notes and experiment logs; the author checked every statement, number and reference.

\bibliographystyle{plainnat}
\bibliography{refs}

\appendix
\section{Reproducibility}
\label{app:repro}

Code, layouts and logs are in the accompanying repository. Per model the whole study is one generated shell chain: pre-tokenise FineWeb-Edu, run the affinity pass, build the eight optimised or ablation layouts and the eight random layouts, then the 23 training runs; every run is skipped if its log is complete, so the chain is resumable. The figures and tables are produced from the logs by one script. Hyper-parameters not stated in \S\ref{sec:setup}: micro-batch 4 with 4 accumulation steps; gradient checkpointing on; calibration and training tokens from the same pre-tokenised file with the first 64 blocks reserved for calibration; random layouts drawn with fixed seeds 1000 to 1007; the optimiser's neuron assignment uses 300 dual-ascent iterations with step decay 0.98 before the fix-up and swap phases.

\section{Earlier runs and the shared-expert variant}

Before the controlled study we ran a 2000-step version of the same recipe on \qwen{} with one optimised and one random layout (perplexity 16.42 and 17.82, KL 0.265 and 0.346 at 2000 steps; the optimised run reached the random run's final perplexity at step 814). A variant in which 10\% of each layer's neurons are replicated on every device, added locally and never broadcast, finished at 16.26 / 0.260 under the same protocol, within noise of the plain optimised layout, so we did not pursue it in the controlled study.

\section{Additional figures: \danube}

\begin{figure}[H]
\centering
\includegraphics[width=\linewidth]{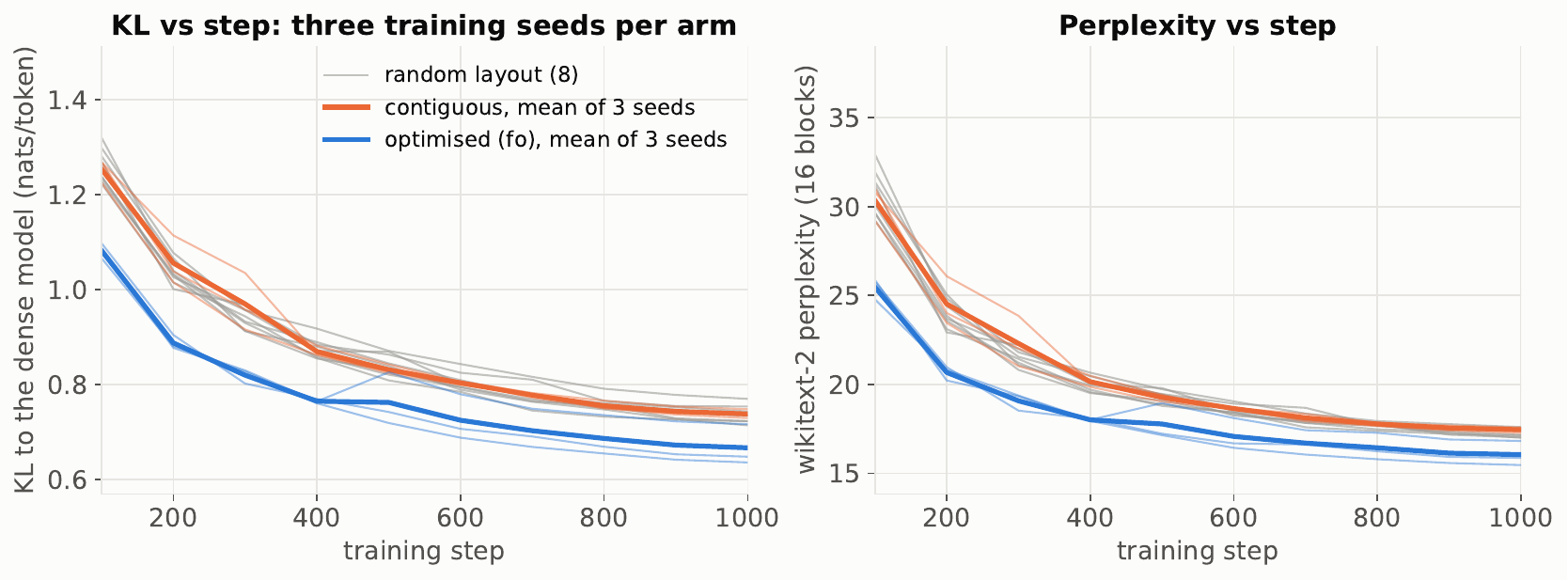}
\caption{\danube{} seed curves, as Figure~\ref{fig:seeds}.}
\label{fig:danube_seeds}
\end{figure}

\begin{figure}[H]
\centering
\includegraphics[width=\linewidth]{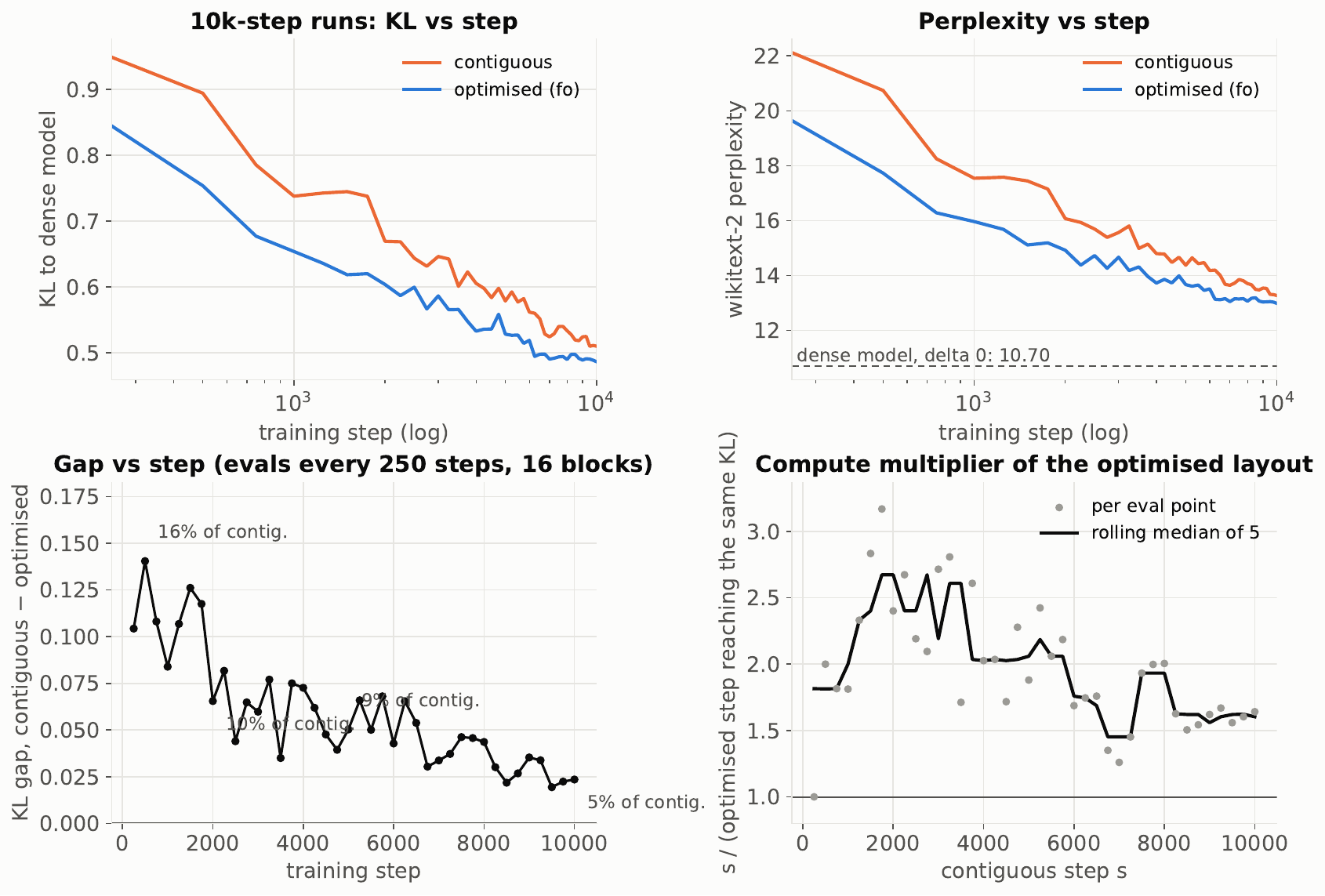}
\caption{\danube{} 10k-step runs, as Figure~\ref{fig:gap}. The relative gap falls from 16\% to 4.6\% but the optimised run is ahead at all 40 evaluations; the multiplier is 1.6 to 2.4$\times$ at the tabulated steps and its rolling median of five stays between 1.45 and 2.7$\times$.}
\label{fig:danube_gap}
\end{figure}

\begin{figure}[H]
\centering
\includegraphics[width=\linewidth]{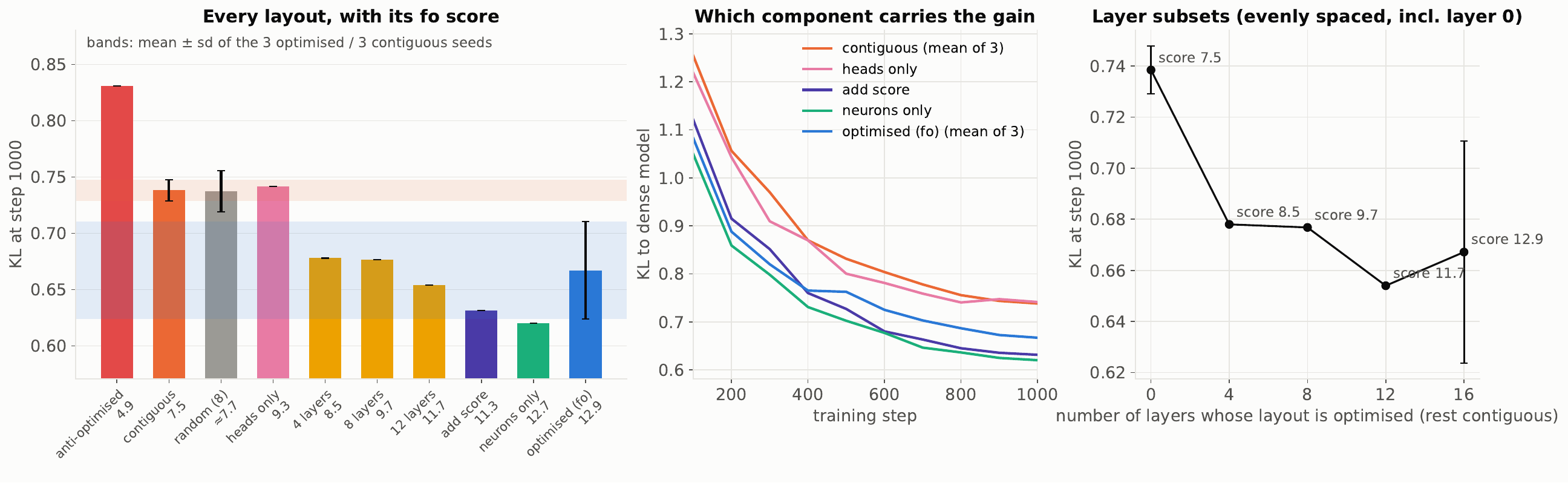}
\caption{\danube{} ablations, as Figure~\ref{fig:ablations}. Neurons-only is the best 1000-step layout; heads-only equals contiguous; four evenly spaced layers give 85\% of the gain.}
\label{fig:danube_ablations}
\end{figure}

\section{Per-layer co-located fractions}

\begin{figure}[H]
\centering
\includegraphics[width=\linewidth]{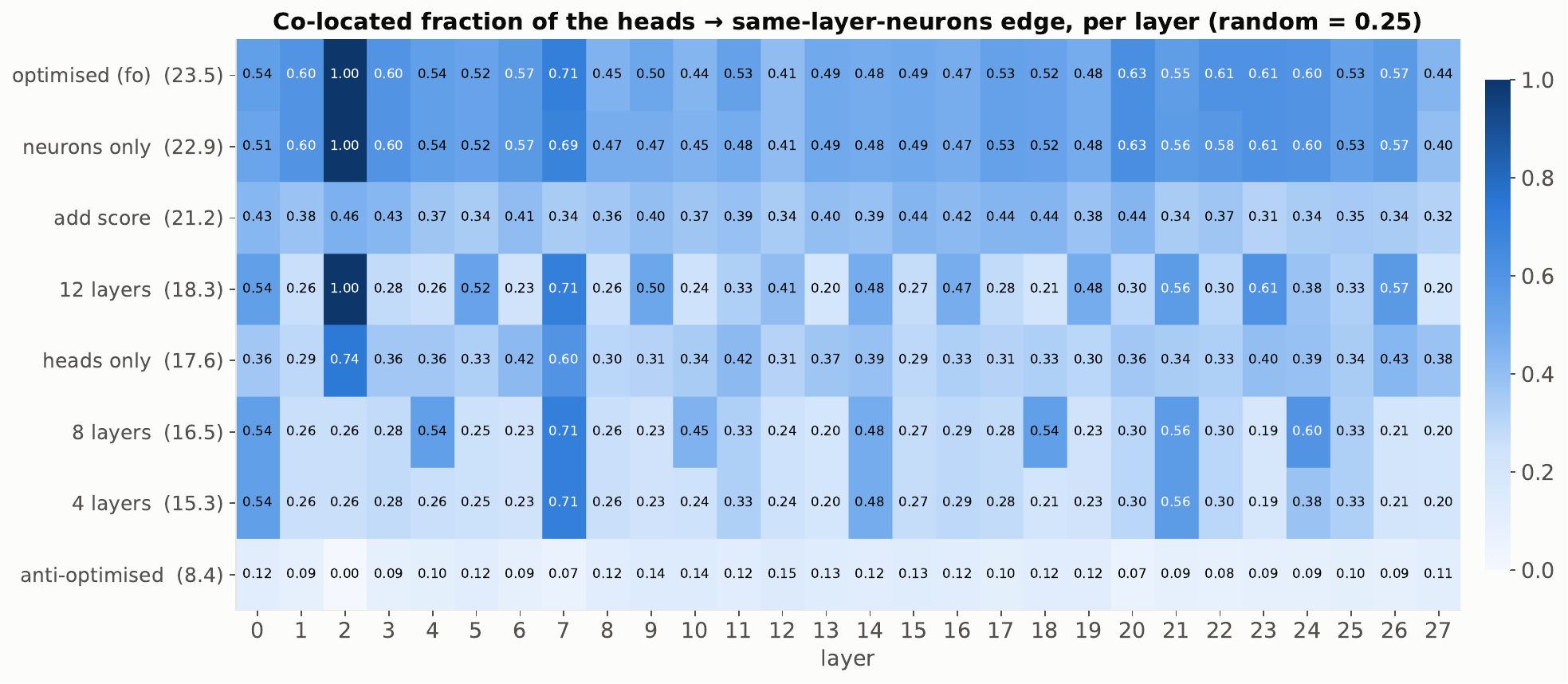}
\caption{\qwen: co-located fraction of the up edge (heads to same-layer neurons) per layer for each optimised layout, random $=0.25$. Layer 2, which hosts the massive-activation circuit of \S\ref{sec:circuit}, is fully separable: the optimiser co-locates 100\% of its affinity, the anti-optimised layout 0\%, and the head sweep alone already reaches 0.74 with contiguous neurons. Eight layers reach 0.6 or more and none is below 0.41.}
\label{fig:qwen_layers}
\end{figure}

\begin{figure}[H]
\centering
\includegraphics[width=\linewidth]{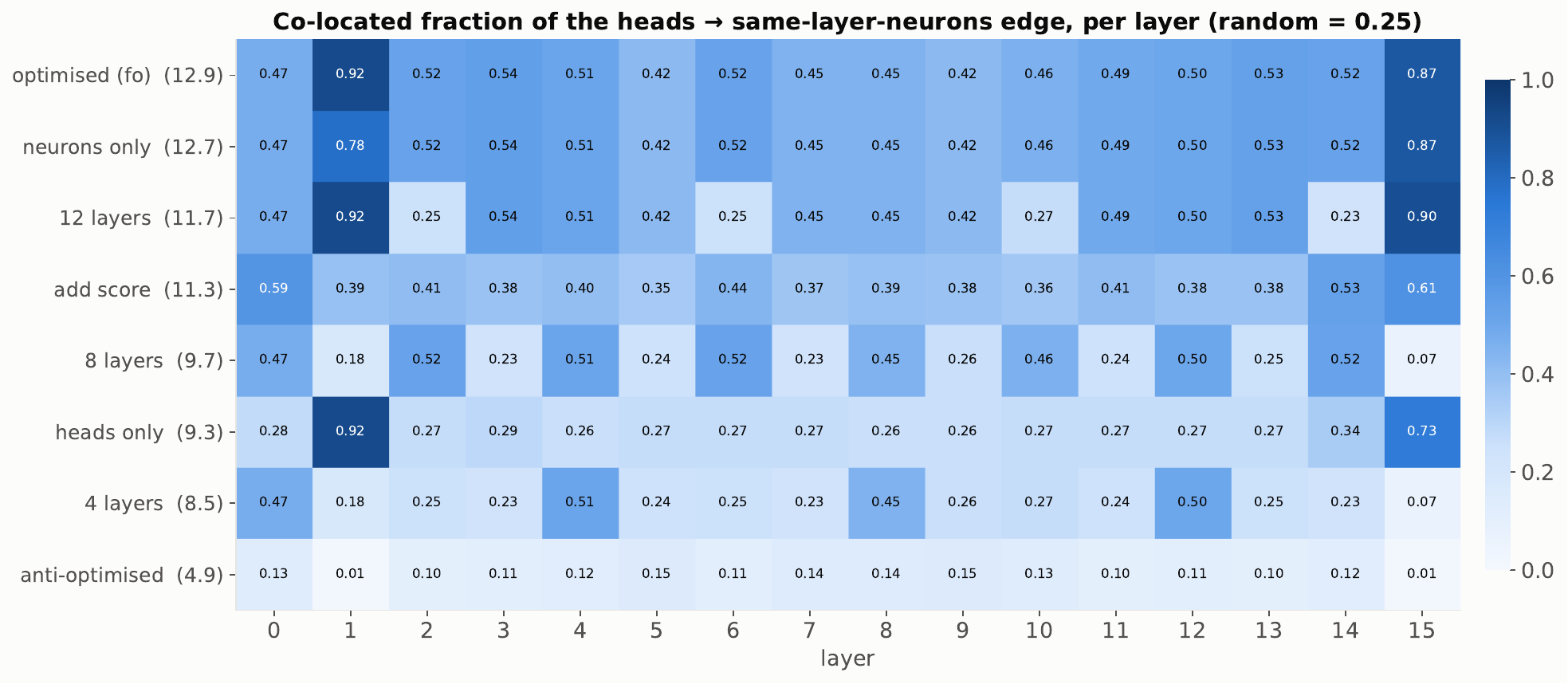}
\caption{\danube: as Figure~\ref{fig:qwen_layers}. Layers 1 and 15 are the only ones above 0.6; layer 1 is the only layer where the head sweep adds to the neuron assignment (0.78 with contiguous heads, 0.92 with optimised heads).}
\label{fig:danube_layers}
\end{figure}

\section{Untrained versus trained}

\begin{figure}[H]
\centering
\includegraphics[width=\linewidth]{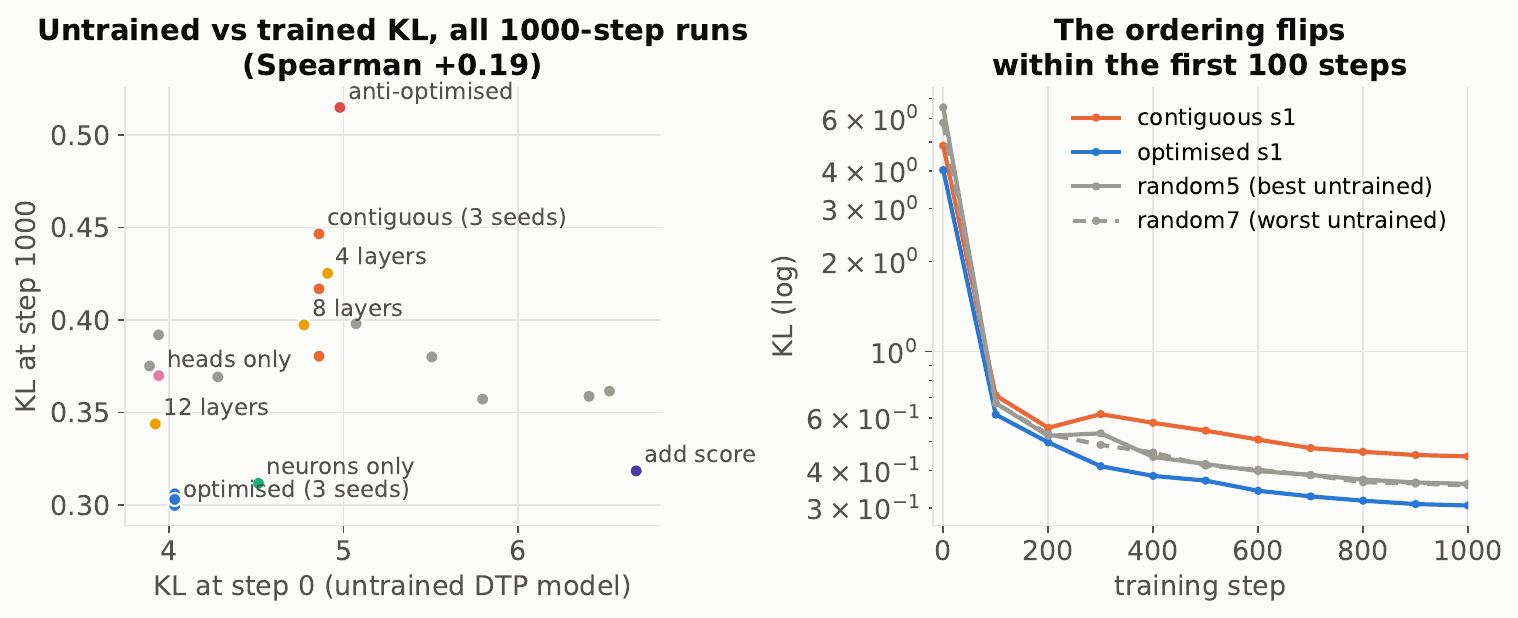}\\[2pt]
\includegraphics[width=\linewidth]{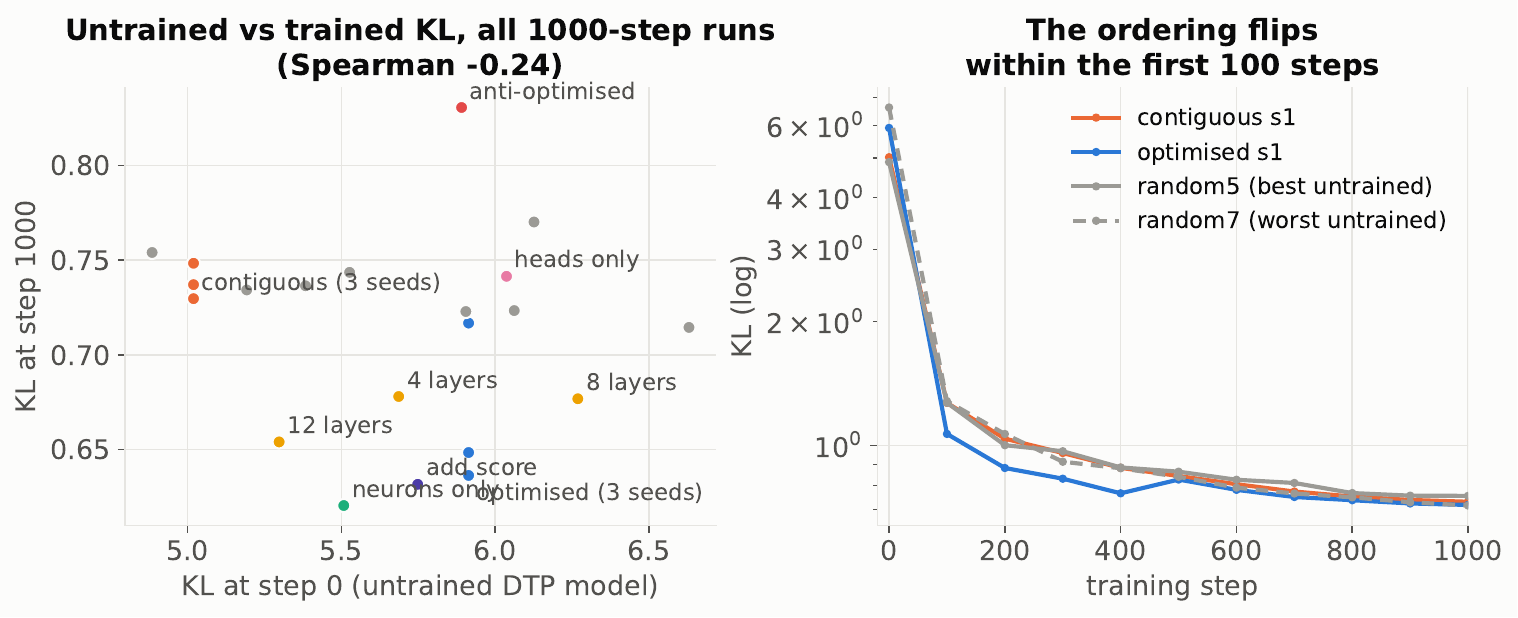}
\caption{Left: KL at step 0 against KL at step 1000 for every 1000-step run (\qwen{} top, \danube{} bottom). Right: full KL trajectories on a log scale for the contiguous and optimised seed-1 runs and the best and worst random layouts untrained. On \danube{} the contiguous layout is the best untrained layout of the seventeen and the optimised one is twelfth; by step 100 the order has flipped and it never flips back.}
\label{fig:untrained}
\end{figure}

\end{document}